\documentclass[conference]{IEEEtran}
\IEEEoverridecommandlockouts
\usepackage{cite}
\usepackage{amsmath,amssymb,amsfonts}
\usepackage{algorithmic}
\usepackage{graphicx}
\usepackage{textcomp}
\usepackage{url}
\usepackage{xcolor}
\def\BibTeX{{\rm B\kern-.05em{\sc i\kern-.025em b}\kern-.08em
    T\kern-.1667em\lower.7ex\hbox{E}\kern-.125emX}}

\newcommand{\Fcal}{\mathcal{F}}

\newcommand*{\mathcolor}{}
\def\mathcolor#1#{\mathcoloraux{#1}}
\newcommand*{\mathcoloraux}[3]{%
  \protect\leavevmode
  \begingroup
    \color#1{#2}#3%
  \endgroup
}

\begin{document}

\title{Rare Failure Prediction via Event Matching for Aerospace Applications\\
\thanks{The research was partially supported by the Russian Foundation for Basic Research grants 16-29-09649 ofi m.}
}

\author{\IEEEauthorblockN{Evgeny Burnaev}
\IEEEauthorblockA{\textit{Skoltech}\\
Moscow, Russia \\
e.burnaev@skoltech.ru}
}

\maketitle

\begin{abstract}
%\textcolor{red}{Abstract}
In this paper, we consider a problem of failure prediction in the context of predictive maintenance applications. We present a new approach for rare failures prediction, based on a general methodology, which takes into account peculiar properties of technical systems. We illustrate the applicability of the method on the real-world test cases from aircraft operations.
\end{abstract}


\section{Prognostic Health Management in Aerospace Industry}
\label{intro}

%Intro and Overview.
%
%\section{Prediction of Failures and }
%\label{aerospace}

The aerospace industry is one of the most heavily regulated industries. With such an emphasis on safety and product quality and with such economic and health consequences due to equipment failures effective maintenance process plays the crucial role in the product success. That is why it is very important to predict and prevent possible failures, reduce repair costs and increase fleet availability while adhering to the rules and procedures set out by the regulatory bodies.

This often leads to the maintenance departments performing more preventative work than it is necessary to increase assurance in equipment reliability, even if this extra precautions do not always provide any additional benefits.

Even more, during certification process of the aerospace equipment specific maintenance policies are being developed. Usually airlines are stick to this policies and do not take any actions to improve them.

In recent years big aircraft manufacturers and airline companies declared that Aircraft Prognostic Health Management (PHM) converts aircraft data into actionable information by leveraging deep engineering knowledge and in-service fleet experience, and provides great possibilities to
\begin{itemize}
\item Determine the operational status of the equipment,
\item Evaluate present condition of the equipment,
\item Detect abnormal conditions in a timely manner,
\item Initiate actions to prevent possible outages, 
\end{itemize}
see, e.g., \cite{boeing, cacao, aircraftPHM, burn2014d, burnPHM}.

The important features of the PHM application are that 
\begin{itemize}
\item Aircraft (A/C) data has a very complex structure:
\begin{itemize}
\item[---] high-dimensional time-series (dimension is usually more than several hundreds);
\item[---] measurement rate could be very high (up to tens of thousands of observations for each flight); at the same time the measurement rate could be different for different parameters; 
\item[---] big volumes of data (a typical size of a historial data sample is measured in terabytes);
\item[---] missing values, non-stationary noise;
\item[---] complex hierarchical structure of a nomenclature of failure types;
\item[---] complex structure and distributed nature of the corresponding data storage;
\end{itemize}
\item failures are rare events with adverse effects; at the same time classical statistical predictive models are ineffective for such events because of their rarity;
\item when predicting failures we have to provide a significant coverage of accurately predicted failures and at the same time have a very low false alarm rate.
\end{itemize}

As a result development of a full support automated system for the early warnings of possible costly faults and failure prediction is a very challenging task. That is why many applications in the field of A/C predictive maintenance are based on simple ``threshold'' monitoring rules capable of detecting only simple faults and having high false alarm rates. However, it is not enough for costly failures anticipation.

E.g. for each of Finnair's eight aircrafts (A330 and A340) during 2012 due to problems with a bleed system about 20 hr. of delays occurred, which costs about 100 euros/minute. Not to pick on Airbus, but the manufacturer's Airman aircraft monitoring system either provided warnings rather late or did not provide  warnings at all \cite{airman}. One of the reasons is a lack of efficient methods for failure prediction. Only after Finnair decided to put its faith in math and asked an engineering company that develops specific mathematical algorithms for improvement of industrial production, to attack the problem, Finnair got reliable service and improved company's air fleet availability.

%\textcolor{red}{PHM papers including for A/C. Nothing works. Why?}

In this work we develop a methodology for building a predictive maintenance policy for complex systems such as aircraft engines. We will demonstrate on examples about real aircraft operations that the developed methodology can efficiently provide failure anticipation and warning monitoring function to decide whether an  operability-related failure is present in the  aircraft before a fault actually occurs.

%\textcolor{red}{Why our approach works? Paper structure?}

The paper has the following structure. In Section \ref{methodology} we describe the developed methodology. In Section \ref{event} we describe the algorithm for event matching being on of the important parts of the proposed approach for failure prediction.
In Section \ref{data} we provide description of the aircraft data. In Sections \ref{engine} and \ref{oil} we describe results of application of the methodology for two use cases. We make conclusions in Section \ref{conclusions}.

\section{General Methodology}
\label{methodology}

Engineering equipment (say, aircraft engine) typically falls into a pre-failure state starting with some minor flaws, e.g. cracks or leaks, that evolve in time and  can lead up to critical failure events such as complete engine destruction. The natural need of the maintenance engineers is to identify these flaws (anomalies) as early as possible and thus try to prevent or even avoid critical events or at least to prepare for the event on time. 

In some cases, based on real-time sensor observations, it is possible to indirectly identify the anomalies in the system behavior related to the minor problems, since the observations being monitored undergo changes in their distributions in response to a change in the environment or, more generally, to changes in certain patterns. Here the development of accurate and reliable mathematical models and tools comes up to the stage. 

In some industries, e.g. aviation, it is crucial to have models and decision-making strategies with a maximum predictive power and a strictly limited false alarm rate. Moreover, it is important to decompose a black-box predictive model to explain an engineer the obtained prediction, which gives her hints on how to act further.

%\textcolor{red}{References? Some smooth transition here}

Let us describe the main steps of the methodology, which grounds on the natural considerations about failure precursors, discussed above.

\textbf{Step 1}. Data filtering and normalization.

\textbf{Step 2}. System decomposition: partition of all the measured parameters into groups, such that the parameters within the group are the most dependent (for example, correlated), but the groups of the parameters are not significantly dependent. Usually such  decomposition corresponds to a physical partitioning of the engineering system into weakly dependent parts corresponding to specific nodes of the engineering system. For the decomposition we can use methods for clustering and community detection in networks \cite{graphclustering}, and graph embedding approaches \cite{AWE}.

\textbf{Step 3}. Detection and classification of various types of anomalies in combinations of observed physical parameters within each of the clustered groups of dependent parameters. The occurrence of an anomaly within the group of dependent parameters indicates a change in the dependencies between these parameters, which in turn means a change in the mode of operation of the corresponding part of the engineering system described by this group of parameters. Thus such anomaly can be a precursor of a future failure of the entire system. Due to the wide variety of data types, it is necessary to use various methods for anomaly detection:
\begin{itemize}
\item some sensor data is represented in time-series format, so we can detect sequences of anomalies in streams of sensor data using  \cite{Multichannel2017,QuasiPeriodic,FBM2016,Degradation2016,ConformalMartingales2017,ConformalAD2015,kNN2017}, and then we can construct ensembles for rare events prediction \cite{EnsemblesDetectors2015,newsmolyakov,AggregationLongTerm, korotin1, korotin2} using detected anomalies and their features as precursors of major failures to optimize specific detection metrics similar to the one used in \cite{Vehicle2017};
\item we can take into account privileged information about the future events, which is accessible during the training stage. Analogous approach, used in \cite{OCSVM2016,OCSVM2018,ModelSelection2015} for anomaly detection with model selection, allowed significant accuracy improvement;
\item historical sensor data has a kind of spatial dimension, since different time-series components correspond to different nodes of the engineering system; thus a graph of dependencies between streams of data, registered by different sensors, can be constructed and modern methods for graph feature learning \cite{AWE,InfluenceSet2018} and panel time-series feature extraction \cite{TDA,TDA2,MF2018} can be applied to enrich the set of input features, used for predictive model construction.
\end{itemize}

\textbf{Step 4}. Associating the detected anomalies with the subsequent (in future flights) failures of specific A/C subsystems on available historic data. At this stage a stream of historic telemetry data is represented by a stream of events (anomalies), detected in each of the groups of  dependent parameters, extracted in Step 2. The hypothesis is that the appearance of particular combinations of anomalies in some of the selected groups of dependent parameters manifests changes in operating modes of specific components of the engineering system, which in turn lead to a failure in the near future. This hypothesis is tested on historical data, which should contain examples of failures that should be predicted. To test the hypothesis we can use methods of imbalanced classification \cite{Imbalance2019,burn2015i}, as well as greedy events matching algorithms. The purpose of these algorithms is to identify subsets of events (anomalies in our case), detection of which reliably predicts some future failures.

\textbf{Step 5}. Built the final model for predicting failures, consisting of several decision rules:
\begin{itemize}
\item[a.] For each group of parameters, extracted in Step 2, we apply the selected set of anomaly detection methods;
\item[b.] For the obtained set of anomalies we check whether there is such subsequence of anomalies among the detected ones, which precedes a failure with a high probability  (according to the historical data);
\item[c.] We note that such model allows us to explain the ``cause'' of a particular forecast:  to identify the input parameters that most affected the forecast. Indeed, a failure is predicted when a certain combination of anomalies is detected; these anomalies correspond to specific groups of parameters. They, in turn, can be associated with specific nodes of the engineering system. 
\end{itemize}

\textbf{Step 6}. Verification of the constructed decision rules based on the cross-validation technique: 
\begin{itemize}
\item[a.] The available historical sample of observations is divided into parts w.r.t. measurements from  different aircrafts; 
\item[b.] The predictive model is trained on all data except the data corresponding to one of the aircrafts;
\item[c.] The accuracy of the trained model prediction is estimated on the data that is not used to train the model;
\item[d.] Actions 6.b and 6.c are repeated the number of times equal to the number of different aircrafts from which the historial data was collected;
\item[e.] The values of model prediction accuracy metrics are aggregated (for example, averaged).
\end{itemize}

This methodology is quite universal and can be applied to various engineering technical systems.

\section{Event Matching}
\label{event}

The goal of an event matching algorihm is to find precursors to failure events of interest in the form of sequences of anomalies. 

Let us describe the proposed event matching algorithm. We denote by 
\begin{itemize}
    \item[---] $(f_t)_{t=1}^T \in \{0, 1\}$ --- alarms (based on anomalies) ($f\in \Fcal$),   $(t^f_j)_{j=1}^{J_f} = \bigl\{t\,:\,f_t = 1 \bigr\}$ --- their firing times, $(y_t)_{t=1}^T\in\{0,1\}$ ---  failures,
    \item[---] $w\geq 1$ --- predictive window, $h\geq 0$ --- horizon (how many moments prior to the onset of an event an alarm is considered anticipatory), $m\geq 0$ --- maintenance action effect delay.
   \end{itemize}
    % h -- how many moments prior to the onset of an event an alarm is considered anticipatory

To construct alarms we apply different anomaly detection algorithms, see Step 3 in Section \ref{methodology}. Then for the alarm $f\in \Fcal$ with respect to the failure event in $[\tau_0, \tau_1)$ we count
\begin{itemize}
      \item true signals: they fire timely, i.e. $t^f_j \in [\tau_0-h-w, \tau_0-h)$;
      \item irrelevant signals due to the event onset or maintenance:
        they fire too early, i.e. $t^f_j \in [\tau_0-h, \tau_1+m)$;
      \item false signals: they fire too early, i.e. $t^f_j < \tau_0-h-w$.
    \end{itemize}

We estimate the predictive performance using the following quantities:
  \begin{itemize}
    \item[---] $K_+$ and $K_-$ --- true and false alarm periods of the target event respectively;
    \item[---] $S^f_+$ and $S^f_-$ --- total number of true and false alarm firings;
    \item[---] $U^f_+$ and $U^f_-$ --- unique true and false alarm firings.
  \end{itemize}

\begin{figure}[t!]
    \centering
    \includegraphics[scale=0.58]{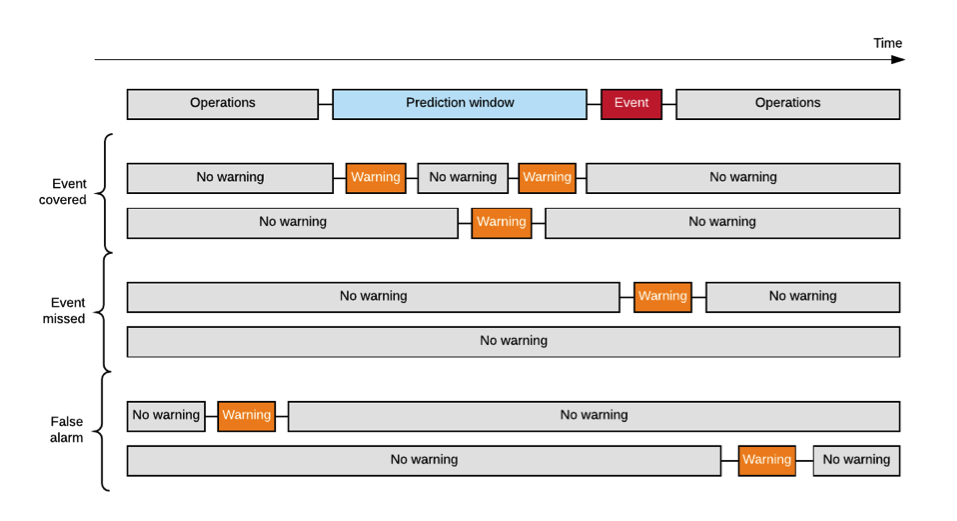}
    \caption{Illustration of accuracy metrics}
    \label{fig1}
  \end{figure}

Thus as efficiency metrics of an early warning system we use
  \begin{itemize}
    \item false alarm rate (precision) $\text{fa}_f = \frac{S^f_-}{K_+} \in [0, +\infty)$;
    \item ratio of covered events (sensitivity) $\text{cf}_f = \frac{U^f_+}{K_+} \in[0, 1]$;
    \item the false alarm ratio $\text{fa/cf}_f = \frac{S^f_-}{U^f_+}$.
  \end{itemize}
Let us note here that we use \textbf{total} false alarm, and
\textbf{unique} true alarm counts. In Fig. \ref{fig1} we provide an illustration, explaning these metrics.

Now let us describe the proposed selection strategy, used to extract sequences of anomalies  (predictive anomalies) that can be utilized as sufficiently accurate precursors of the failure of interest. 

We use two approaches, namely, hard filter and soft filter:
    \begin{itemize}
    \item[1.] First, we perform \textbf{t-test}: we consider some alarm $f$ to be promising for the prediction if $U^f_+ > 1$ and the p-value of the hypothesis $p^f_+ = p^f_-$ vs. the hypothesis $p^f_+ > p^f_-$ is below the significance level $\alpha$ with $p^f_\epsilon = \frac{S^f_\epsilon}{K_\epsilon}$, $\epsilon\in \{+, -\}$;
    \item[2.] \textbf{Hard filter}: we select the alarm $f$ to be used to predict a failure $y$ whenever $U^f_+ \geq \theta$ and $U^f_- = 0$ for a
    threshold $\theta$ controlling the support of the hypothesis, i.e. we define the implication $f\implies y$;
    \item[3.] \textbf{Soft filter}: we select the alarm $f$ to be used to predict the failure $y$ if $\text{fa/cf}_f \leq \theta$ for a threshold  $\theta$ controlling the false-to-covered ratio.
  \end{itemize}
  
We generate final alarm signals in the following way. For each target event $y$ we get pairs of predictive anomalies
  $\mathcal{P} \subseteq \{ \{a_1, a_2\} \, : \, a_1, a_2 \in \mathcal{F}\}$. Here $A\in \mathcal{P}$ if and only if $y_t$ is considered to be  predictable by the signal $A_t = \wedge_{a_t\in A} a_t$. In some cases we consider not pairs of predictive anomalies but triples. Alarm signals for the event $y$ are synthesized with
\begin{equation}
    S_t = \bigvee_{A_t\in \mathcal{P}} A_t, \label{eqeq12}
\end{equation}
since pooling predictive anomalies with low false alarm rate increases the  chances of successfully anticipating an event. Alarm signals in \eqref{eqeq12} play a role of precursors for the failure $y$ of interest.

Finally, we perform alarm signals synthesis:
  \begin{itemize}
    \item[1.] Group parameters with respect to their semantics, or dependence graph. As a dependence measure we use a simple Pearson correlation, or more complex non-linear measures like mutual information;
    \item[2.] For each group of parameters we use anomaly detection algorithms to detect anomalies:
    \begin{itemize}
      \item The most typical anomaly detection algorithms are based on manifold modeling approaches \cite{RobotLocalization2017,ConformalDR,MLR2018,DRreg}; yet another approach could be to construct a surrogate model \cite{GTApprox2016,Ensembles2013,HDA2013,MFGP2015,MFGP2017} in order to approximate dependencies between the observed parameters and then detect anomalies based on a predictive error with a non-parametric confidence measure \cite{ConformalKRR2016,VovkConformal2014} as the diagnostic indicator;
      \item In a linear case we can use the low rank linear PCA reconstruction error \cite{PCA2015} as the diagnostic time-series; 
      \item Observations with errors, exceeding 90\%-95\% empirical quantile, are considered as anomalies.
    \end{itemize}
    \item[3.] Use the event matching algorithm to find pairs/triplets of predictive anomalies with adequate coverage and false alarm rates.
  \end{itemize}
  
  \section{Aircraft data}
  \label{data}
  
\begin{figure}[t!]
    \centering
    \includegraphics[scale=0.19]{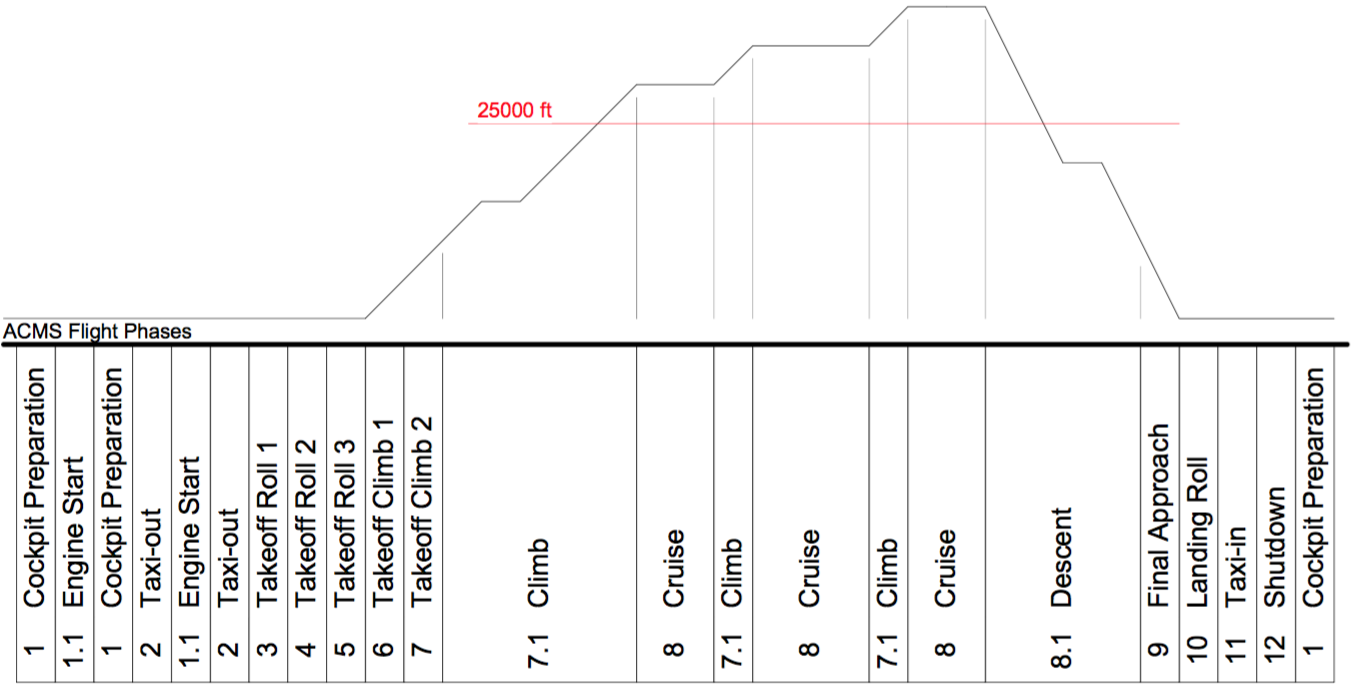}
    \caption{Flight phases}
    \label{fig2}
  \end{figure}

We test our methodology using telemetry of $32$ %\textbf{Airbus A380} 
aircrafts. The data includes:
    \begin{itemize}
    \item[1.] Multiple telemetry snapshots taken only under certain conditions,
    \item[2.] Engine related ACMS reports --- ``Takeoff''(4), ``Climb''(3), and ``Cruise''(1,2), including parameters
    \begin{itemize}
      \item EGTT: Exhaust Gas Temperature trimmed;
      \item OPU: Engine oil pressure;
      \item QDMCNT: Amount of captured abrasive particles in the oil flow;
      \item Nacelle/turbine vibration, HP/LP turbine exit pressure, Exit thrust, $\ldots$;
    \end{itemize}
    \item[---] rep. $01-04$ have $322$, $406$, $318$, and $339$ parameters resp.;
    \item[---] span a year and a half of operations, $\geq 400$ flights per year.
  \end{itemize}
This data corresponds to the aircraft flight phases, depicted in Fig. \ref{fig2}.

The Engine Central Maintenance System logs track various events:
  \begin{itemize}
    \item[---] Fault codes: low-level indicators of unusual conditions in a circuit component or its subsystem;
    \item[---] Warnings: high-level indication of failures, time-outs, etc.: 
    \begin{itemize}
      \item \texttt{7100w4X0}: Engine stall (shutdown);
      \item \texttt{7400wXX0}: Engine igniter A/B fault;
      \item \texttt{7830wXX0}: Engine reverser thruster inhibited/unlocked.
      \item \texttt{4962W0X0}: Auxiliary Power Unit Fault;
    \end{itemize}
    \item[---] ATA / JASC code grouping (Joint Aircraft System/Component (JASC) Code Tables, Air Transport Association of America (ATA)): \begin{itemize}
      \item 7XXX -- Turbine engine \begin{itemize}
        \item 73XX -- engine fuel and control;
        \item 74XX -- ignition,
        \item 77XX -- oil filter clogging.
      \end{itemize}
    \end{itemize}
    \item[---] Data is higly imbalanced: most flights experience no warnings.
  \end{itemize}

\section{Engine Shutdown prediction} % (fold)
\label{engine}

This use case concentrates on the unexpected engine shutdown failure. Engine shutdown is
an in-flight failure and it is reported as CMS messages with codes \texttt{7100W310},
\texttt{7100W320}, \texttt{7100W330}, and \texttt{7100W340}. In these codes, the last
but one digit corresponds to the number of engine failed.

The goal is to predict future occurrences of engine shutdown critical failure using
historical data from ACMS and CMS reports of 32 aircrafts. For this study we used the set of reports (1--4); here it is important to distinguish
between different flight phases. Only engine related features were employed (refer
to Step 2 in Sec.~\ref{methodology} and Sec. \ref{data}).

Fortunately, unexpected engine shutdowns are exceptionally rare: during only $5$
flights out of $34777$ shutdown events were encountered, and for only $4$ aircrafts out of $32$.

We set the parameters of the early warning system (see Sec.~\ref{event})
as follows: horizon $h=0$, window $w=20$ flights, and the maintenance effect $m=0$.

Due to the extreme rarity of analyzed events, it was impossible to employ the automatic
anomaly detection (see Step 3 in Sec. ~\ref{methodology}) and the automated predictive
alarms selection (see Sec. \ref{event}). Also due to severity of the analyzed failure, the restrictions on the false alarm rate were less stringent.

In case of low-frequency events the probability estimates, used in Sec. \ref{event}, are unreliable.
Thus we can not apply the method of Sec.~\ref{event} straightforwardly and optimal features had to be hand-picked taking into account the prediction quality metric. As a result we acquired at phase ``7.1'' the following features:  ``R3::FFDP\_B1x'', ``R3::P25\_H1x'', and
``R3::ZTFEFA\_K1x'' from report ``3''.

\begin{figure*}[t!]
  \centering
    \includegraphics[scale=0.45]{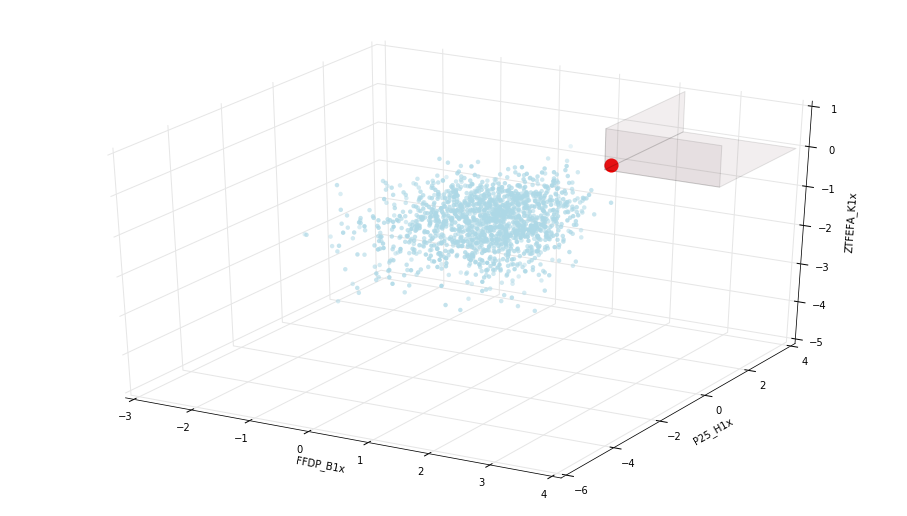}
  \caption{The engine shutdown event: snapshot values of alarm signals for the aircraft is
  highlighted in pale red, and abnormality thresholds, as given in table~\ref{tab:engshut_table}, are depicted in pale  grey}
  \label{fig:engshut_plot1}
\end{figure*}
%\subsection{Anomaly Extraction}

In order to perform Anomaly Extraction in this case the lower dimensional linear data manifold was learnt over
the sample of flights which were sufficiently separated from shutdown events. This sample, the so called ``normal'' regime, is consists of
\begin{enumerate}
  \item all historical flights of an aircraft, which never encountered a shutdown
  event;
  \item all flights of aircrafts, which did encounter the event, that are at least
  $50$ moments before or $30$ moments after the shutdowns.
\end{enumerate}
This sample represents the normal regime of coupling of parameters between engines,
which is why it is used to estimate a rank-1 approximation of the parameter group's
intrinsic linear manifold. The abnormality scores are calculated as reconstruction
error of the recorded within group measurements %(sec.~\ref{ssub:anomaly_extraction})
of the complete sample of flights of all aircrafts, see Figure \ref{fig:engshut_plot1}.

\begin{table}[t!]
\center
  \begin{tabular}{c|cccc}
  Phase & Feature (report 3) & Threshold \\ \hline
  71 & FFDP\_B1x   & 8.69175  \\
     & P25\_H1x    & 2.096015 \\
     & ZTFEFA\_K1x & 1.01905  \\
  \end{tabular}
\caption{Abnormality thresholds for engine shitdown event ``7100W330''}
\label{tab:engshut_table}
\end{table}

\begin{figure}[t!]
  \centering
    \includegraphics[scale=0.38]{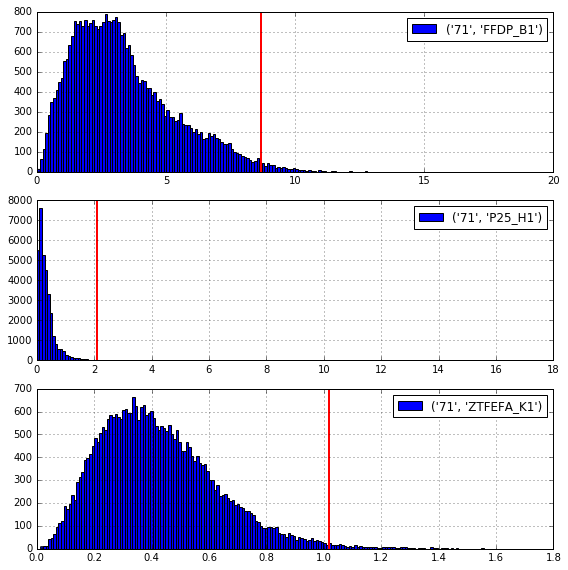}
  \caption{Histograms of abnormality scores' distributions with picked thresholds}
  \label{fig:engshut_plot2}
\end{figure}

The threshold for the abnormality score in a certain parameter group is based on a mix
of $\{95\%, 99\%\}$ empirical quantiles, obtained via manual search and optimization
process (see Tab.~\ref{tab:engshut_table} and Figure \ref{fig:engshut_plot2}).

\section{Oil filter clogging prediction} 
\label{oil}

The problem is to predict oil filter clogging in order to optimize maintenance of an oil filter (minimization of the number of inspections and cost of supplies). To do this we have to automatically extract parameters  tied with the oil filter clogging event and to construct a model for prediction of the filter clogging using observed data. The prediction problem is imbalanced as the number of failures (filter cloggings) is small compared to the number of examples of the normal regime.

We considered a subset of $10$ parameters (that could be the most relevant to the failure) and applied the automatic parameter selection methodology together with the imbalanced classification \cite{burn2015i,Imbalance2019}. The successive elimination of parameters finally left only several parameters, including OPU (Oil pressure) etc., which occurred to be the most related to the failure. 

At the next step we divided the  data into the train/test sets and applied the failure prediction procedure that accounts for the autoregression depth of parameters values used when making predictions; we tune the depth values based on the prediction error. It occurred that the history up to $3$ flights back provides the most accurate predictions.

\begin{figure}[t!]
  \centering
    \includegraphics[width = 0.5\textwidth]{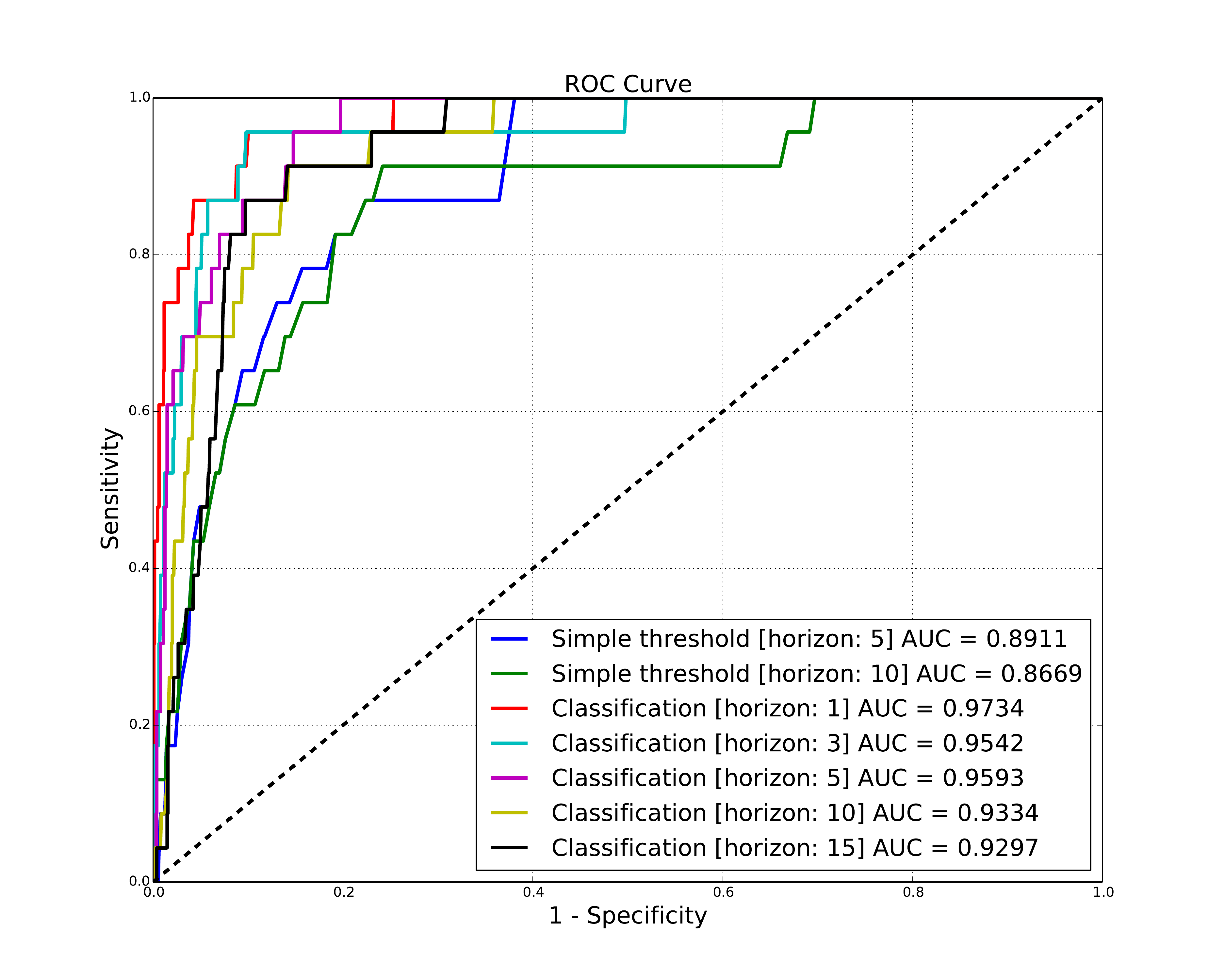}
   \caption{Low Oil Pressure: ROC curves}\label{fig:leop_roc}.
    \end{figure}

\begin{figure}
  \centering
    \includegraphics[width = 0.5\textwidth]{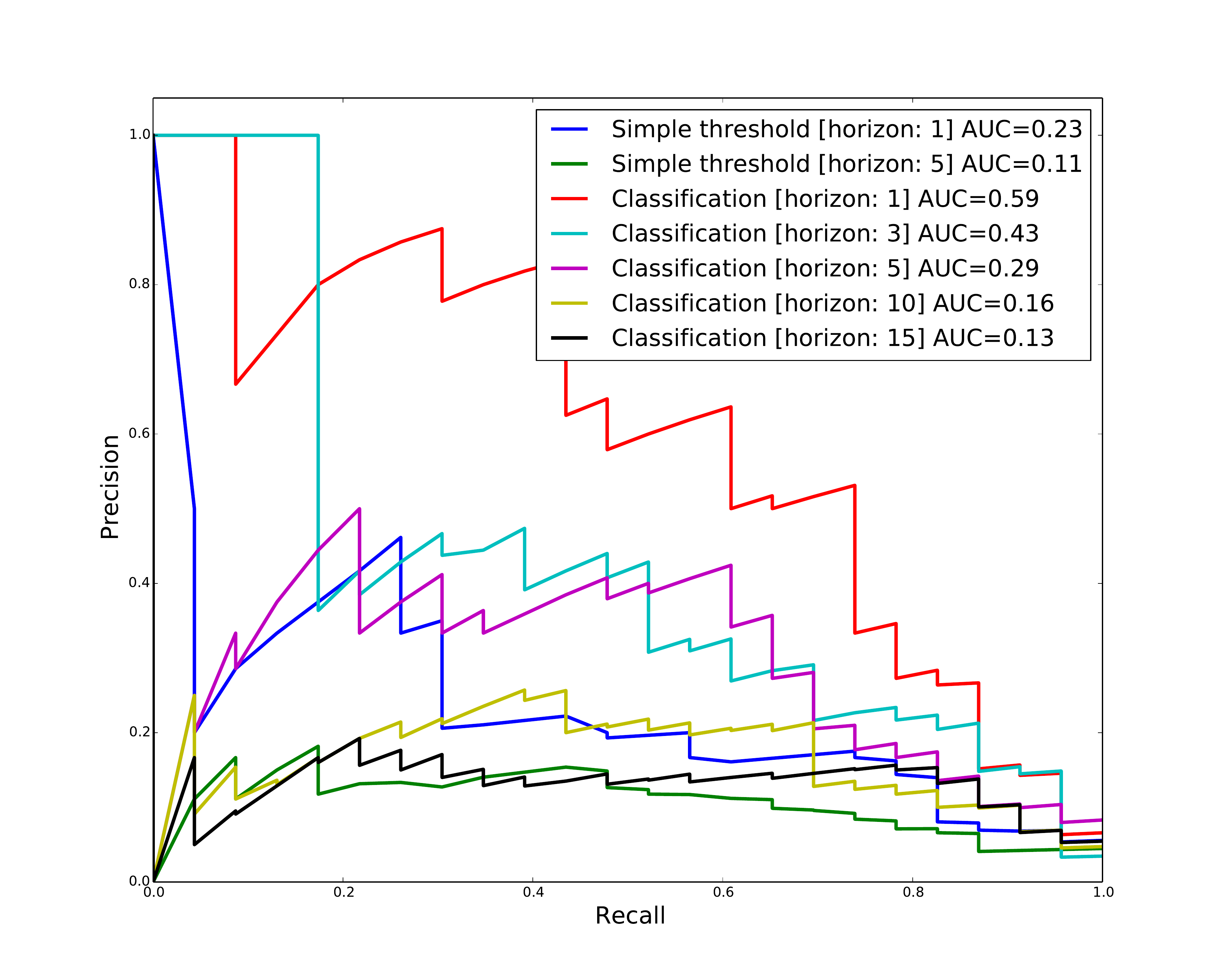}
   \caption{Low Oil Pressure: Precision-Recall curves}\label{fig:leop_pr}.
\end{figure}

To understand benefits of the proposed approach we compared it with the simple thresholding algorithm, which predicts failures by comparing the level of OPU with some predefined threshold. The results for the different horizon of prediction along with the simple thresholding prediction based on OPU level are presented in Figures~\ref{fig:leop_roc} and \ref{fig:leop_pr}. 
The proposed approach exposes better metrics compared to the simple thresholding, thus allowing to predict failure events earlier with the same level of false alarms.

\begin{figure}
  \centering
    \includegraphics[width = 0.5\textwidth]{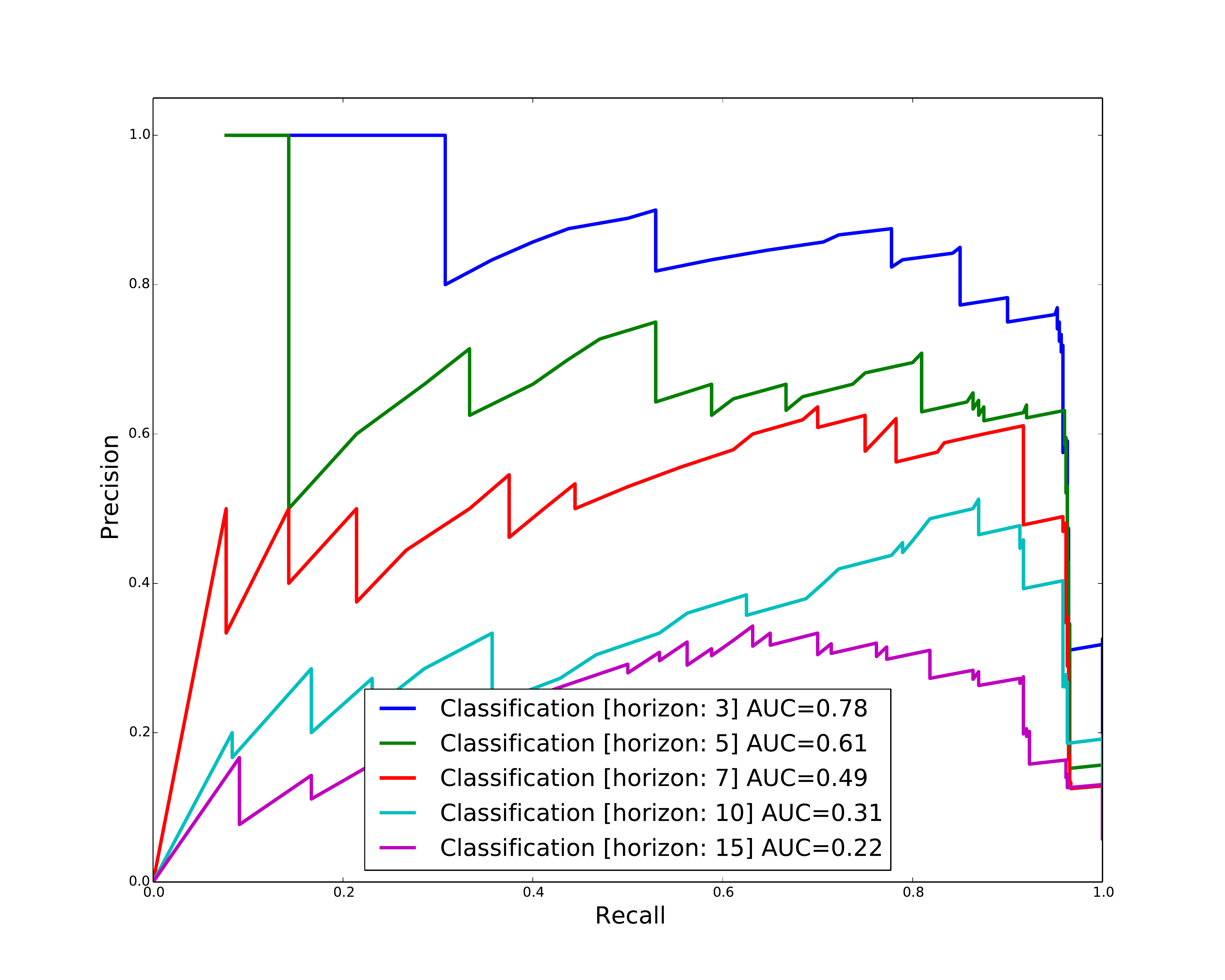}
   \caption{Low Oil Pressure: Precision-Recall curves with $\pm 2$ flights tolerance}\label{fig:leop_pr_withtol}.
\end{figure}

In practice the prediction of a failure in the vicinity of a real one is also acceptable and not considered as false alarm. The results for failures predicted in the range of $\pm 2$ flights near the real faults are presented in Figure \ref{fig:leop_pr_withtol}.

\begin{figure}
  \centering
    \includegraphics[width=0.48\textwidth,]{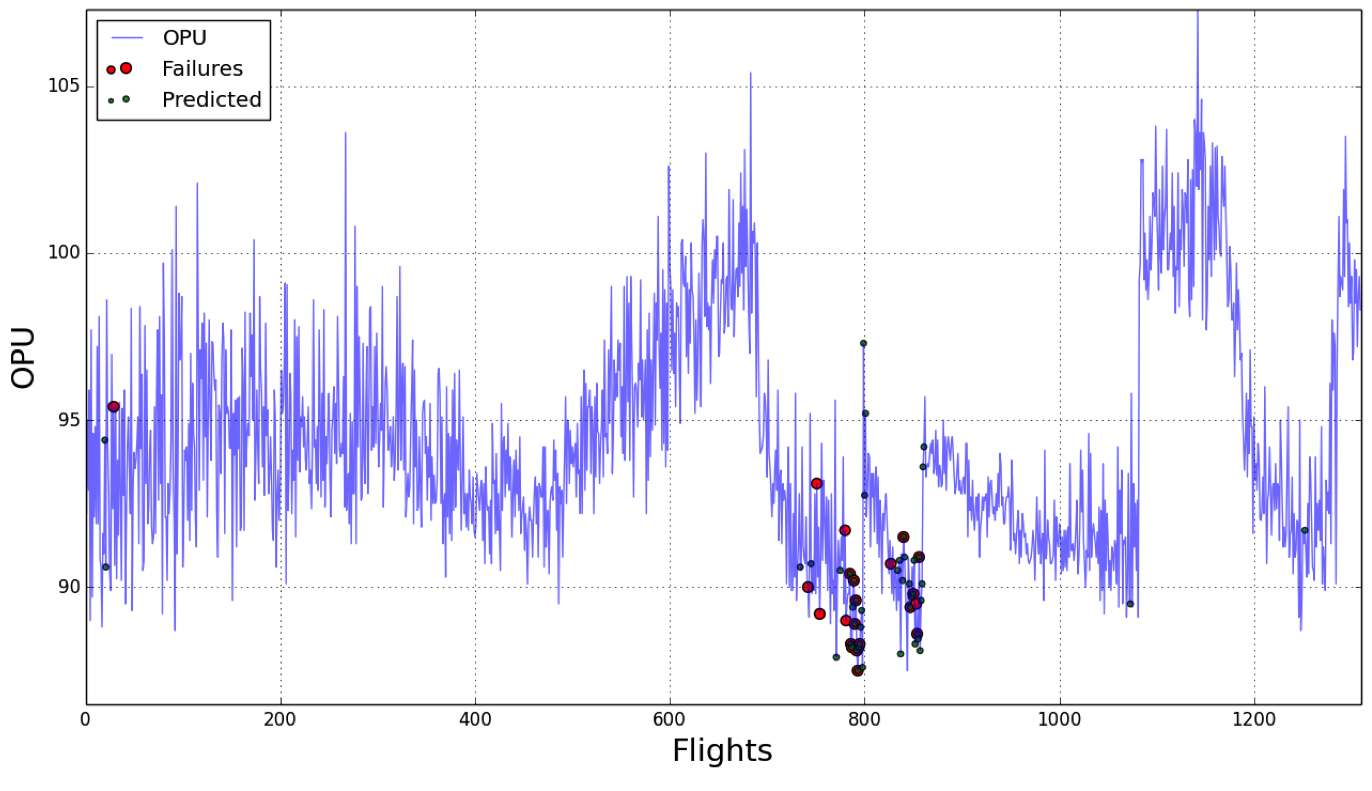}
   \caption{Low Oil Pressure: Prediction in time (confidence threshold $\nu=0.6$)}\label{fig:leop_5fw-0.6}.
\end{figure}

The particular predictions in time for the classifier confidence threshold $\nu=0.6$ are shown in Figure \ref{fig:leop_5fw-0.6}.

\section{Conclusions}
\label{conclusions}

In this paper we proposed a data-driven approach to the rare failure prediction problem. Thanks to the elaborated approach we were able to predict some possible aircraft equipment failures. The next steps would be to automate the methodology, as it still requires a manual choice of some hyperparameters (various thresholds), as well as to consider more use cases to expand the proposed methodology.

\section*{Acknowledgment}

E. Burnaev would like to thank I. Nazarov and P. Erofeev for a help with data processing, and company Datadavnce llc. for the problem statement.

%\section*{References}

%\bibliographystyle{splncs04}

%\bibliographystyle{plain}
%\bibliography{Burnaev_PHM_v2.bib}

\bibliographystyle{IEEEtran}
\bibliography{IEEEabrv,Burnaev_PHM_v2}

\begin{thebibliography}{10}
\providecommand{\url}[1]{#1}
\csname url@samestyle\endcsname
\providecommand{\newblock}{\relax}
\providecommand{\bibinfo}[2]{#2}
\providecommand{\BIBentrySTDinterwordspacing}{\spaceskip=0pt\relax}
\providecommand{\BIBentryALTinterwordstretchfactor}{4}
\providecommand{\BIBentryALTinterwordspacing}{\spaceskip=\fontdimen2\font plus
\BIBentryALTinterwordstretchfactor\fontdimen3\font minus
  \fontdimen4\font\relax}
\providecommand{\BIBforeignlanguage}[2]{{%
\expandafter\ifx\csname l@#1\endcsname\relax
\typeout{** WARNING: IEEEtran.bst: No hyphenation pattern has been}%
\typeout{** loaded for the language `#1'. Using the pattern for}%
\typeout{** the default language instead.}%
\else
\language=\csname l@#1\endcsname
\fi
#2}}
\providecommand{\BIBdecl}{\relax}
\BIBdecl

\bibitem{boeing}
\BIBentryALTinterwordspacing
\emph{Maintenance optimization. Airplane health management}, 2015. [Online].
  Available:
  \url{http://www.boeing.com/resources/boeingdotcom/commercial/services/assets/brochure/airplanehealthmanagement.pdf}
\BIBentrySTDinterwordspacing

\bibitem{cacao}
T.~Shen, F.~Wan, W.~Cui, and B.~Son, ``Application of prognostic and health
  management technology on aircraft fuel system,'' in \emph{IEEE Proceedings of
  2010 Prognostics and System Health Management Conference}.\hskip 1em plus
  0.5em minus 0.4em\relax Macao: IEEE, 12--14 Jan 2010, pp. 1--7.

\bibitem{aircraftPHM}
J.~Dai and H.~Wang, ``Evolution of aircraft maintenance and logistics based on
  prognostic and health management technology,'' in \emph{Lecture Notes in
  Electrical Engineering. Proceedings of the First Symposium on Aviation
  Maintenance and Management-Volume II}, vol. 297.\hskip 1em plus 0.5em minus
  0.4em\relax Springer, 2014, pp. 665--672.

\bibitem{burn2014d}
S.~Alestra, C.~Bordry, C.~Brand, E.~Burnaev, P.~Erofeev, A.~Papanov, and
  C.~Silveira-Freixo, ``Application of rare event anticipation techniques to
  aircraft health management,'' \emph{Advanced Materials Research}, vol. 1016,
  pp. 413--417, 2014.

\bibitem{burnPHM}
------, ``Rare event anticipation and degradation trending for aircraft
  predictive maintenance,'' in \emph{Proceedings of the joint WCCM – ECCM –
  ECFD 2014 Congress, 20-25 July, Barcelona, Spain}, 2014, pp. 1--12.

\bibitem{airman}
L.~Tegtmeier, ``Math and maintenance,'' \emph{Aviation Week and Space
  Technology}, vol. 174, no.~39, 2012.

\bibitem{graphclustering}
B.~Saha, A.~Mandal, S.~Tripathy, and D.~Mukherjee, ``Complex networks,
  communities and clustering: {A} survey,'' \emph{CoRR}, vol. abs/1503.06277,
  2015.

\bibitem{AWE}
S.~Ivanov and E.~Burnaev, ``Anonymous walk embeddings,'' in \emph{Proc. of the
  35th ICML}, vol.~80.\hskip 1em plus 0.5em minus 0.4em\relax PMLR, 2018, pp.
  2186--2195.

\bibitem{Multichannel2017}
E.~V. Burnaev and G.~K. Golubev, ``On one problem in multichannel signal
  detection,'' \emph{Problems of Information Transmission}, vol.~53, no.~4, pp.
  368--380, 2017.

\bibitem{QuasiPeriodic}
A.~Artemov, E.~Burnaev, and A.~Lokot, ``Nonparametric decomposition of
  quasi-periodic time series for change-point detection,'' in \emph{Proc.
  SPIE}, vol. 9875, 2015, pp. 9875--9875--5.

\bibitem{FBM2016}
A.~Artemov and E.~Burnaev, ``Optimal estimation of a signal perturbed by a
  fractional brownian noise,'' \emph{Theory of Probability \& Its
  Applications}, vol.~60, no.~1, pp. 126--134, 2016.

\bibitem{Degradation2016}
------, ``Detecting performance degradation of software-intensive systems in
  the presence of trends and long-range dependence,'' in \emph{2016 IEEE 16th
  International Conference on Data Mining Workshops (ICDMW)}, 2016, pp. 29--36.

\bibitem{ConformalMartingales2017}
D.~Volkhonskiy, E.~Burnaev, I.~Nouretdinov, A.~Gammerman, and V.~Vovk,
  ``Inductive conformal martingales for change-point detection,'' in
  \emph{Proceedings of the Sixth Workshop on Conformal and Probabilistic
  Prediction and Applications}, vol.~60.\hskip 1em plus 0.5em minus 0.4em\relax
  PMLR, 2017, pp. 132--153.

\bibitem{ConformalAD2015}
A.~Safin and E.~Burnaev, ``Conformal kernel expected similarity for anomaly
  detection in time-series data,'' \emph{Advances in Systems Science and
  Applications}, vol.~17, no.~3, pp. 22--33, 2017.

\bibitem{kNN2017}
V.~Ishimtsev, A.~Bernstein, E.~Burnaev, and I.~Nazarov, ``Conformal k-nn
  anomaly detector for univariate data streams,'' in \emph{Proceedings of the
  Sixth Workshop on Conformal and Probabilistic Prediction and Applications},
  vol.~60.\hskip 1em plus 0.5em minus 0.4em\relax PMLR, 2017, pp. 213--227.

\bibitem{EnsemblesDetectors2015}
A.~Artemov and E.~Burnaev, ``Ensembles of detectors for online detection of
  transient changes,'' in \emph{Proc. SPIE}, vol. 9875, 2015, pp. 9875 -- 9875
  -- 5.

\bibitem{newsmolyakov}
D.~{Smolyakov}, N.~{Sviridenko}, V.~{Ishimtsev}, E.~{Burikov}, and
  E.~{Burnaev}, ``{Learning Ensembles of Anomaly Detectors on Synthetic
  Data},'' \emph{arXiv e-prints}, vol. abs/1905.07892, 2019.

\bibitem{AggregationLongTerm}
A.~Korotin, V.~V'yugin, and E.~Burnaev, ``Aggregating strategies for long-term
  forecasting,'' in \emph{Proceedings of the Seventh Workshop on Conformal and
  Probabilistic Prediction and Applications}, vol.~91.\hskip 1em plus 0.5em
  minus 0.4em\relax PMLR, 2018, pp. 63--82.

\bibitem{korotin1}
A.~{Korotin}, V.~{V'yugin}, and E.~{Burnaev}, ``{Adaptive Hedging under Delayed
  Feedback},'' \emph{arXiv e-prints}, vol. abs/1902.10433, 2019.

\bibitem{korotin2}
------, ``{Long-Term Online Smoothing Prediction Using Expert Advice},''
  \emph{arXiv e-prints}, vol. abs/1711.03194, 2017.

\bibitem{Vehicle2017}
E.~Burnaev, I.~Koptelov, G.~Novikov, and T.~Khanipov, ``Automatic construction
  of a recurrent neural network based classifier for vehicle passage
  detection,'' in \emph{Proc. SPIE}, vol. 10341, 2017, pp. 10\,341--10\,341--6.

\bibitem{OCSVM2016}
E.~Burnaev and D.~Smolyakov, ``One-class svm with privileged information and
  its application to malware detection,'' in \emph{2016 IEEE 16th International
  Conference on Data Mining Workshops (ICDMW)}, 2016, pp. 273--280.

\bibitem{OCSVM2018}
D.~Smolyakov, N.~Sviridenko, E.~Burikov, and E.~Burnaev, ``Anomaly pattern
  recognition with privileged information for sensor fault detection,'' in
  \emph{Artificial Neural Networks in Pattern Recognition}.\hskip 1em plus
  0.5em minus 0.4em\relax Springer, 2018, pp. 320--332.

\bibitem{ModelSelection2015}
E.~Burnaev, P.~Erofeev, and D.~Smolyakov, ``Model selection for anomaly
  detection,'' in \emph{Proc. SPIE}, vol. 9875, 2015, pp. 9875 -- 9875 -- 6.

\bibitem{InfluenceSet2018}
S.~Ivanov, N.~Durasov, and E.~Burnaev, ``Learning node embeddings for influence
  set completion,'' in \emph{Proc. of IEEE International Conference on Data
  Mining Workshops (ICDMW)}, 2018, pp. 1034--1037.

\bibitem{TDA}
R.~Rivera, P.~Pilyugina, A.~Pletnev, I.~Maksimov, W.~Wyz, and E.~Burnaev,
  ``Topological data analysis of time series data for b2b customer relationshop
  management,'' in \emph{Proc. of Industrial Marketing and Purchasing Group
  Conference (IMP19)}, ser. The IMP Journal, 2019.

\bibitem{TDA2}
R.~{Rivera-Castro}, I.~{Nazarov}, Y.~{Xiang}, A.~{Pletneev}, I.~{Maksimov}, and
  E.~{Burnaev}, ``{Demand forecasting techniques for build-to-order lean
  manufacturing supply chains},'' \emph{arXiv e-prints}, vol. abs/1905.07902,
  2019.

\bibitem{MF2018}
R.~Rivera, I.~Nazarov, and E.~Burnaev, ``Towards forecast techniques for
  business analysts of large commercial data sets using matrix factorization
  methods,'' \emph{Journal of Physics: Conference Series}, vol. 1117, no.~1, p.
  012010, 2018.

\bibitem{Imbalance2019}
D.~Smolyakov, A.~Korotin, P.~Erofeev, A.~Papanov, and E.~Burnaev,
  ``Meta-learning for resampling recommendation systems,'' in \emph{Proc. SPIE
  11041, Eleventh International Conference on Machine Vision (ICMV 2018),
  110411S (15 March 2019)}, 2019.

\bibitem{burn2015i}
E.~Burnaev, P.~Erofeev, and A.~Papanov, ``Influence of resampling on accuracy
  of imbalanced classification,'' in \emph{Proc. SPIE}, vol. 9875, 2015, pp.
  9875--9875--5.

\bibitem{RobotLocalization2017}
A.~Kuleshov, A.~Bernstein, E.~Burnaev, and Y.~Yanovich, ``Machine learning in
  appearance-based robot self-localization,'' in \emph{2017 16th IEEE
  International Conference on Machine Learning and Applications (ICMLA)}, 2017,
  pp. 106--112.

\bibitem{ConformalDR}
A.~Kuleshov, A.~Bernstein, and E.~Burnaev, ``Conformal prediction in manifold
  learning,'' in \emph{Proceedings of the Seventh Workshop on Conformal and
  Probabilistic Prediction and Applications}, vol.~91.\hskip 1em plus 0.5em
  minus 0.4em\relax PMLR, 2018, pp. 234--253.

\bibitem{MLR2018}
------, ``Manifold learning regression with non-stationary kernels,'' in
  \emph{Artificial Neural Networks in Pattern Recognition}.\hskip 1em plus
  0.5em minus 0.4em\relax Springer, 2018, pp. 152--164.

\bibitem{DRreg}
------, ``Kernel regression on manifold valued data,'' in \emph{Proceedings of
  IEEE 5th International Conference on Data Science and Advanced Analytics},
  2018, pp. 120--129.

\bibitem{GTApprox2016}
M.~Belyaev, E.~Burnaev, E.~Kapushev, M.~Panov, P.~Prikhodko, D.~Vetrov, and
  D.~Yarotsky, ``Gtapprox: Surrogate modeling for industrial design,''
  \emph{Advances in Engineering Software}, vol. 102, pp. 29--39, 2016.

\bibitem{Ensembles2013}
E.~V. Burnaev and P.~V. Prikhod'ko, ``On a method for constructing ensembles of
  regression models,'' \emph{Automation and Remote Control}, vol.~74, no.~10,
  pp. 1630--1644, 2013.

\bibitem{HDA2013}
M.~G. Belyaev and E.~V. Burnaev, ``Approximation of a multidimensional
  dependency based on a linear expansion in a dictionary of parametric
  functions,'' \emph{Informatics and its Applications}, vol.~7, no.~3, pp.
  114--125, 2013.

\bibitem{MFGP2015}
E.~Burnaev and A.~Zaytsev, ``Surrogate modeling of multifidelity data for large
  samples,'' \emph{Journal of Communications Technology and Electronics},
  vol.~60, no.~12, pp. 1348--1355, 2015.

\bibitem{MFGP2017}
A.~Zaytsev and E.~Burnaev, ``Large scale variable fidelity surrogate
  modeling,'' \emph{Annals of Mathematics and Artificial Intelligence},
  vol.~81, no.~1, pp. 167--186, 2017.

\bibitem{ConformalKRR2016}
E.~Burnaev and I.~Nazarov, ``Conformalized kernel ridge regression,'' in
  \emph{2016 15th IEEE International Conference on Machine Learning and
  Applications (ICMLA)}, 2016, pp. 45--52.

\bibitem{VovkConformal2014}
E.~Burnaev and V.~Vovk, ``Efficiency of conformalized ridge regression,'' in
  \emph{Proceedings of The 27th Conference on Learning Theory}, vol.~35.\hskip
  1em plus 0.5em minus 0.4em\relax PMLR, 2014, pp. 605--622.

\bibitem{PCA2015}
E.~Burnaev and S.~Chernova, ``On an iterative algorithm for calculating
  weighted principal components,'' \emph{Journal of Communications Technology
  and Electronics}, vol.~60, no.~6, pp. 619--624, 2015.

\end{thebibliography}

%\begin{thebibliography}{00}

%\end{thebibliography}

\end{document}